\documentclass[runningheads]{llncs}
\usepackage{graphicx}
\usepackage{amsmath,amssymb} % define this before the line numbering.
\usepackage{lineno}
\usepackage{color}
\begin{document}
\renewcommand\thelinenumber{\color[rgb]{0.2,0.5,0.8}\normalfont\sffamily\scriptsize\arabic{linenumber}\color[rgb]{0,0,0}}
%\renewcommand\makeLineNumber {\hss\thelinenumber\ \hspace{6mm} \rlap{\hskip\textwidth\ \hspace{6.5mm}\thelinenumber}}
%\linenumbers
\pagestyle{headings}
\mainmatter
\def\IVS16SubNumber{***}  % Insert your submission number here

\title{A Large-scale Distributed Video Parsing and Evaluation Platform} % Replace with your title

%\titlerunning{IVS-16 submission ID \IVS16SubNumber}

%\authorrunning{IVS-16 submission ID \IVS16SubNumber}

\author{Kai Yu, Yang Zhou, Da Li, Zhang Zhang, Kaiqi Huang}
\institute{
Center for Research on Intelligent Perception and Computing,
Institute of Automation,
Chinese Academy of Sciences,
China
}

\maketitle

\begin{abstract}

Visual surveillance systems have become one of the largest data sources of Big Visual Data in real world. However, existing systems for video analysis still lack the ability to handle the problems of scalability, expansibility and error-prone, though great advances have been achieved in a number of visual recognition tasks and surveillance applications, e.g., pedestrian/vehicle detection, people/vehicle counting. Moreover, few algorithms explore the specific values/characteristics in large-scale surveillance videos. To address these problems in large-scale video analysis, we develop a scalable video parsing and evaluation platform through combining some advanced techniques for Big Data processing, including Spark Streaming, Kafka and Hadoop Distributed Filesystem (HDFS). Also, a Web User Interface is designed in the system, to collect users' degrees of satisfaction on the recognition tasks so as to evaluate the performance of the whole system. Furthermore, the highly extensible platform running on the long-term surveillance videos makes it possible to develop more intelligent incremental algorithms to enhance the performance of various visual recognition tasks.

\end{abstract}

\section{Introduction}

Intelligent visual surveillance (IVS) has long been one of the most important applications of computer vision technologies. Intelligent surveillance video analysis is a demand-increasing research topic along with the raising of consciousness of public security and explosive increase of deployment of surveillance devices. In the past decades, most researchers aimed to solve separate visual tasks, e.g., background modeling \cite{franccois1999adaptive,rodriguez2016incremental,zhang2014background}, object detection \cite{nadimi2004physical,tripathi2016context,oreifej2013simultaneous}, motion tracking \cite{xu1999adaptive,foxlin2014motion}, person re-identification \cite{hamdoun2008person,wang2016person} and attribute recognition \cite{fukui2016robust,deng2014pedestrian}, which is because early researchers \cite{wang2003recent} considered the complexity of whole surveillance task, thus divided the whole system into several separate steps with a divide-and-conquer strategy. And one ordinary IVS system is simple a fixed execution flow of these sub-tasks. However, the era of Big Data raises new challenges for IVS systems. First, to explore values in large-scale visual surveillance video, it is urged to solve the problem of scalability and error-prone of large-scale video data processing. Second, to discover correlations in various visual information, it need solve the problem of expansibility and flexibility of deploying new visual analysis modules. Third, the system could be improved adaptively and incrementally in its running lifetime, thus the users's feedbacks should be collected to optimize the models for different visual tasks. Unfortunately, in previous work, few efforts were devoted to these problems.

In this work, to address these problems, we present a novel Large-scale Video Parsing and Evaluation Platform (LaS-VPE Platform) based on some advanced techniques for Big Data processing, including Spark Streaming, Kafka and HDFS \cite{shvachko2010hadoop}. This platform can run easily and flexibly on distributed clusters, making full use of large-scale computation resources. High-level abstraction of vision tasks and usage of Spark Streaming and Kafka make it possible to add or replace any algorithm modules at any time and robust to faults, thus easy to maintain and computation-resources-saving. The high flexibility of the system also enables users to specify their own execution plan. Also the well-designed platform and Web UI make it easy for both developers to extend the system and users to operate on the system.

The main contributions of this paper are listed below:
\begin{enumerate}
    \item We propose a detailed and integrated solution for surveillance scene parsing and performance evaluation with large-scale video data.
    \item We solve some technical issues in adopting and integrating Spark Streaming and Kafka in large-scale video processing.
    \item The implementation of this platform is an open-source project, and we will share it in GitHub, so anyone can make use of it while referring to this paper.
\end{enumerate}

\section{Relating works}

There has been researches on IVS systems which show some common interests with our work. The VisFlow \cite{luvisflow} system also combines machine vision with big data analysis, featuring high accuracy and efficiency. It can compute execution plans for different vision queries, by building a relational dataflow system. Compared to this work, the LaS-VPE platform does not provide optimization for execution plan, but instead enables users to easily create their own plans with Web UI. Parameters and execution order can be specified in any valid form on every query, making it highly customizable. The optimization work is left to users or done by future extern modules.

H. Qi et al. proposed the Visual Turing Test system \cite{qi2015restricted} for deep understanding in long-term and multi-camera captured videos. It presents a well-designed system for video-evaluation utilizing scene-centered representation and story-line based queries. However, this work does not focus on system efficiency. Our system emphasizes less on visual algorithm evaluation concepts, but spend more effort on improving the feasibility of evaluation on massive video data by distributed computing techniques and flexible system design.

\section{Platform design}

In this section, we describe the design of our LaS-VPE platform. The platform is powered by Spark Streaming on YARN and Kafka. For better understanding of the mechanism of this platform, readers are recommended to first read through some introductions and have a rough understanding of mechanisms and terms of Spark Streaming and Kafka.

\subsection{Platform Framework}

The LaS-VPE platform is divided into several modules. Each module is responsible for one kind of system affairs, like meta-data saving, extern message handling, and execution of different versions of different visual recognition algorithms. Each module runs permanently, unless the administrator manually terminates it. Communication among different modules is powered by Kafka, which is a high-throughput, distributed, publish-subscribe messaging system.

The LaS-VPE platform also possesses a Web UI. Users can create a certain task in it, and command the tasks to flow through the modules in a specified graph. A task thus has specified input and execution route, and can end with several outputs, including saving meta-data to databases or hard disks and responding user queries in the Web UI.

\subsection{Task Flow and Communication}

To address the problem in traditional IVS systems that only a few preset execution plans are available and to enhance the expansibility and flexibility of the system, we enable our modules to be executed in user-defined plans. Each execution corresponding to a vision query is called a task. Each task can specify the order of visual recognition sub-modules in a form of flow graph of modules. For simplicity, a flow graph must be directed acyclic. One module may exist more than once in a flow graph, if it needs to be executed more than once, enabling finite execution circles to be dissembled and operated in one task.

\begin{figure}
  \centering
  \includegraphics[width=80mm]{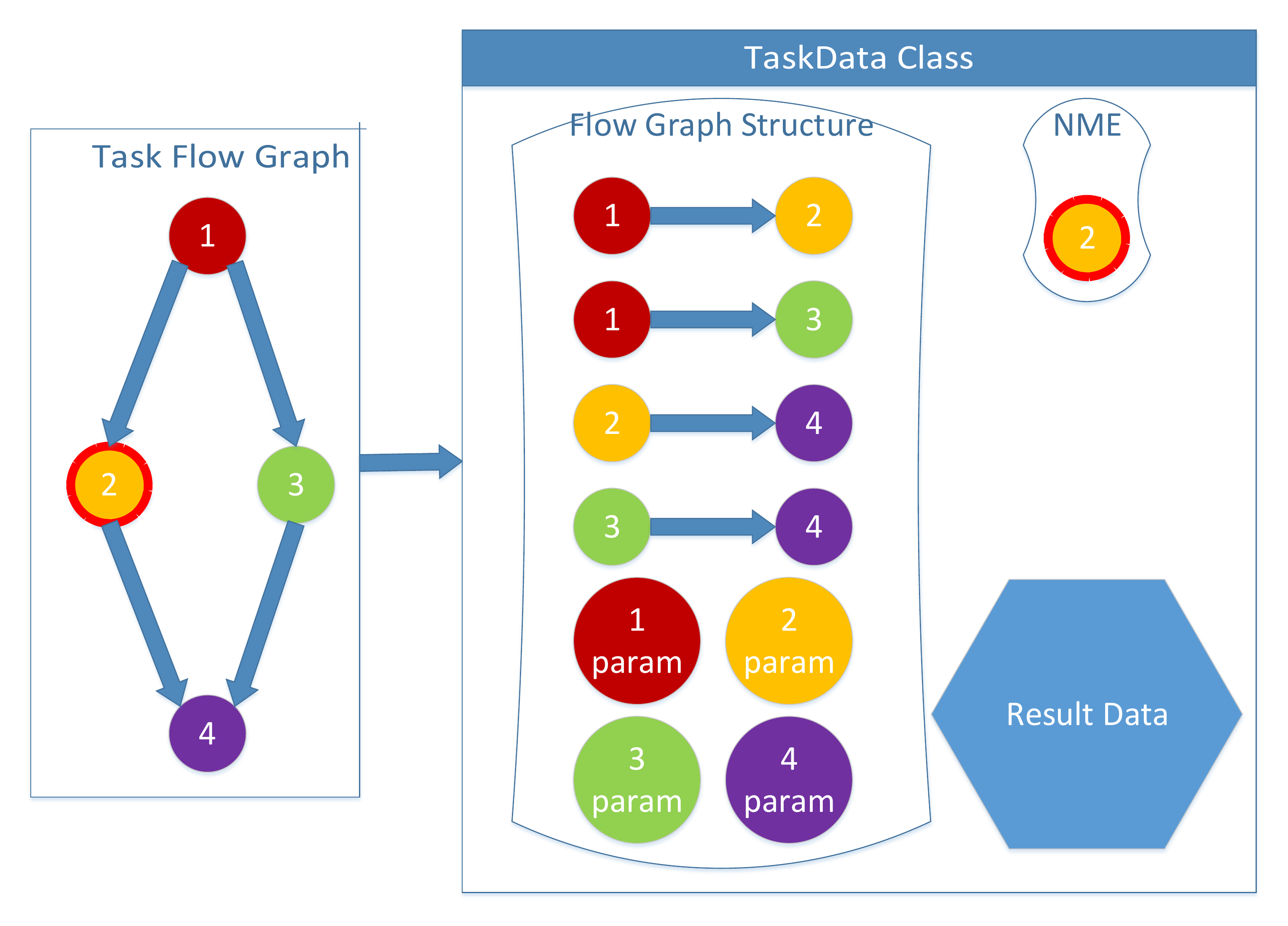}\\
  \caption{This picture illustrates how to represent a logical task flow graph in a TaskData class. The flow graph structure is an adjacency list of the graph. The NME field   specified the next module to execute. In this picture, this module corresponds to the node numbered 2. The result data field's content is not considered here. It will be analyzed by the consuming part of the module.}
  \label{fig:TaskData}
\end{figure}

Tasks pass among modules as a flow of Kafka messages. A Kafka message contains a key field and a value field. The key field records a Universal Unique Identifier (UUID) assigned to each task, which is generated along with the task generation in the Web UI. The value field records a serialized byte array of a special class TaskData, which is illustrated in Figure \ref{fig:TaskData}. It contains three fields: result data from the sender module, identifier of next module to execute (NME) and the flow graph structure of the task. A flow graph structure is then described by two parts: nodes and links, where each node specifies a module to be executed as well as some parameters and extra data for that single execution. Each directed link indicates that after the execution of the module specified by the head node, its results should be sent to the module specified by the tail node, and that module shall be executed some time afterwards.

Since the graph is a directed acyclic graph, it is able to be topological-sorted. Each time a module receives a message, the NME corresponds to the module itself, so the module can find itself in the graph according to this field. If the module takes more than one input node and some inputs currently have not arrived yet, the module cannot be immediately executed, and the current message sent to this module will be cached temporarily. Multiple pieces of cached messages will be merged and accumulated in the module. When the input requirement is satisfied, the module starts an execution with the accumulated data.

\subsection{Kafka-based Inter-module Communication}

We use Kafka for communication between modules, so as to decouple the system and enhance the flexibility and extensibility of the system. By doing so, modules need not be written and compiled together and thus become isolated. When one module dies due to some exception or deliberate termination, it does not affect the others, and the input data to this module will be cached by the Kafka, so when this module recovers or is replaced by a new version of it, these input data can be correctly consumed and processed. Also, the destinations of output can be specified by the task flow graph described above, so it need not modify and recompile a module if we want it to output to some alternative destinations.

Kafka topics can be divided into groups for different type of messages, such as topics for pedestrian tracks and topics for pedestrian attributes. However, topics are not shared among different vision modules. Each module owns one or more topics corresponding to its input data types. For example, if Module \emph{M1} and \emph{M2} both take in pedestrian attributes and tracks as input, \emph{M1} owns two topics named \emph{M1-Pedestrian-Attribute} and \emph{M1-Pedestrian-Track} respectively, while \emph{M2} owns another two topics named \emph{M2-Pedestrian-Attribute} and \emph{M2-Pedestrian-Track} respectively. If some modules have produced some attributes and is commanded to send to both \emph{M1} and \emph{M2}, the Kafka producers inside these modules need to send the attributes to both \emph{M1-Pedestrian-Attribute} and \emph{M2-Pedestrian-Attribute} message queues. In this way, the task flow graph actually does not specifies which module to output to, but instead which topic to output to, since a topic belongs to exactly one module.

\subsection{Spark Streaming and YARN}

For large-scale video data processing, clusters are preferred for high through-out computation. We choose Spark Streaming to enable distributed computation since it guarantees realtime processing, and use YARN to manage the cluster. We view each module as a Spark Streaming application. Applications are submitted to YARN by the SparkLauncher class programmatically, then run permanently. Multiple applications may run simultaneously and independently on YARN. It is easy to terminate applications using the Web UI of YARN. This means we can choose only part of the modules in the platform to run, and terminate any of them whenever we want, thus saving computation resources.

In the LaS-VPE platform, a streaming context usually consists of three stages: Kafka message receiving stage, message re-organizing stage (optional) and message processing stage. The Kafka message receiving stage receive messages from Kafka then transfers them into Discretized Streams (DStreams), which abstractly represents a continuous stream of data. In each call of the final message processing stage, a certain extern algorithm, such as a tracking algorithm, is run simultaneously on multiple workers to process the data delivered to them. The results are then output in various forms, like using a KafkaProducer class to send it to Kafka, or using a FileSystem class to send it to HDFS, etc.

\subsection{Web UI Design and Evaluation}

The LaS-VPE platform provides a Web UI for generating tasks, monitoring the applications in the platform, and querying the visual recognition results. The task generating UI allows users to easily create a flow graph of jobs within each task. The UI server is responsible to transform the graph into the data form mentioned above and send it to the processing modules, then listen to execution results.

The web page for querying results does not directly communicate with the processing modules. Since we force each processing module to save their meta-data and results on HDFS or databases, the query-solving server seeks results at these locations. This makes results endurable and easy to access outside the processing cluster. Also, a feedback field is provided in the result displaying pages, varying according to the type of results, to allow users to simply mark their satisfaction of the results or provide detailed revision on the results they see. For example, in Re-ID applications, results are displayed in a form of ranked candidate photos that are predicted to be the most similar to the target, so users can give feedback by simply selecting the ground-truth ones, and satisfaction as well as supervision information can be both inferred. These feedbacks are stored into a database and can be used for future semi-supervised or supervised incremental training of all the algorithms used for the tasks.

\begin{figure}
  \centering
  \includegraphics[width=107mm]{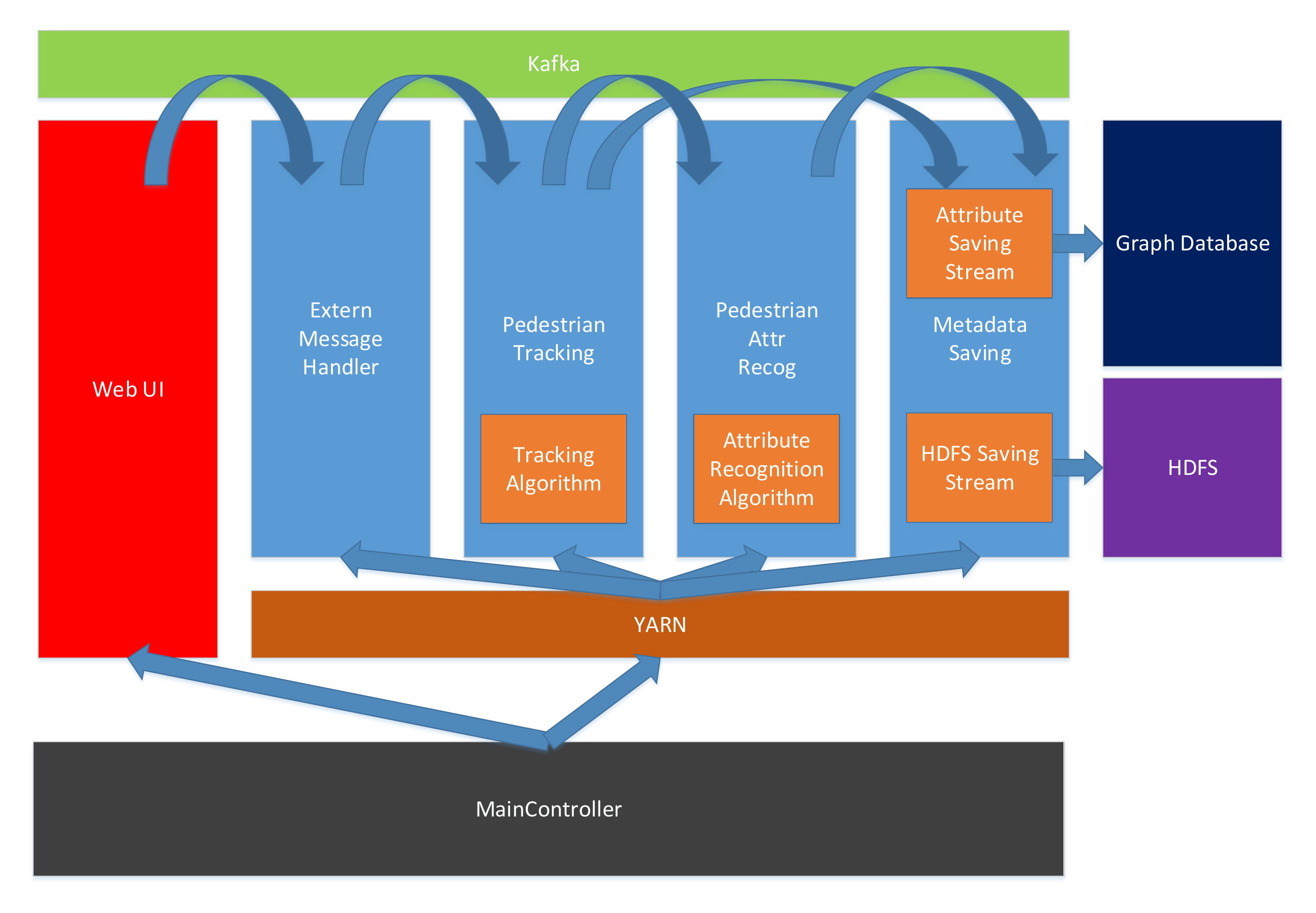}\\
  \caption{This picture shows a sample framework. Modules are submitted to YARN, and task flows start from extern web UI. Results are saved in the meta-data saving module.}
  \label{fig:SystemDesign}
\end{figure}

Combining these features, one can specify various combinations of algorithms and configurations of each vision algorithm modules, then evaluate them on any database representing a particular application scene, so it is easy and low-costing to find out the best settings for any new applications scenes, thus exploiting the ability of video parsing algorithms in maximum extent. The whole system is illustrated in Figure \ref{fig:SystemDesign}.

\section{Conclusions}

In this paper, we proposed a novel Video Parsing and Evaluation Platform to solve problems existing in ordinary video surveillance systems like low flexibility and extensibility and lack of user feedback collecting functions. The VPE-Platform can run robustly on distributed clusters, support highly customized job flow and easy maintaining, and collect user feedback for incremental training. Some experiments will be added in the future to show the design validity and high performance of the platform.

\section{Acknowledgements}

This work was supported by the National Natural Science Foundation of China under Grants 61473290, and the international partnership program of Chinese Academy of Sciences, grant No. 173211KYSB20160008.

\bibliographystyle{splncs}
\bibliography{egbib}

\end{document}